\pdfoutput=1

\documentclass[11pt]{article}

\usepackage[preprint]{acl}

\usepackage{times}
\usepackage{latexsym}
\usepackage{graphicx}
\usepackage{subfigure}
\usepackage{amssymb,amsthm,dsfont,mathrsfs}
\usepackage{amsmath}
\usepackage{setspace}
\usepackage{multirow}
\usepackage{helvet}
\usepackage{courier}
\usepackage{bm}
\usepackage{mdwlist}
\usepackage{booktabs}
\usepackage{algorithm}
\usepackage{algorithmic}
\usepackage{url}
\usepackage{colortbl}
\usepackage{enumerate}
\usepackage{enumitem}
\usepackage{subcaption}
\usepackage{arydshln}
\usepackage[textsize=tiny]{todonotes}

\usepackage[T1]{fontenc}

\usepackage[utf8]{inputenc}

\usepackage{microtype}

\usepackage{inconsolata}

\usepackage{xspace}
\newcommand{\model}{\textsc{LongHeads}\xspace}

\title{\model: Multi-Head Attention is Secretly a Long Context Processor}

\author{  
Yi Lu\textsuperscript{\rm 1}\thanks{\ \ Equal contribution.}, 
Xin Zhou\textsuperscript{\rm 1}\footnotemark[1], 
Wei He\textsuperscript{\rm 1},
Jun Zhao\textsuperscript{\rm 1},
\\
\textbf{Tao Ji\textsuperscript{\rm 1}\thanks{\ \  Corresponding authors.}, 
Tao Gui\textsuperscript{\rm 2}\footnotemark[2], 
Qi Zhang\textsuperscript{\rm 1}\footnotemark[2], 
Xuanjing Huang\textsuperscript{\rm 1,3}}
\\
  {$^1$School of Computer Science, Fudan University, Shanghai, China} \\
  {$^2$ Institute of Modern Languages and Linguistics, Fudan University, Shanghai, China} \\
  {$^3$ International Human Phenome Institutes, Shanghai, China} \\
  \texttt{{yilu23@m.fudan.edu.cn, \{xzhou20,taoji,tgui,qz\}@fudan.edu.cn}}
  }

\begin{document}
\maketitle

\begin{abstract}
Large language models (LLMs) have achieved impressive performance in numerous domains but often struggle to process lengthy inputs effectively and efficiently due to limited length generalization and attention’s quadratic computational demands.
Many sought to mitigate this by restricting the attention window within the pre-trained length. However, these methods introduce new issues such as ignoring the middle context and requiring additional training.
To address these problems, we propose \textbf{\model}, a \emph{training-free} framework that enhances LLM’s long context ability by unlocking multi-head attention’s untapped potential.
Instead of allowing each head to attend to the full sentence, which struggles with generalizing to longer sequences due to out-of-distribution (OOD) issues, we allow each head to process in-distribution length by selecting and attending to important context chunks.
To this end, we propose a chunk selection strategy that relies on the inherent correlation between the query and the key representations, efficiently distributing context chunks to different heads.
In this way, \textbf{each head ensures it can effectively process attended tokens within the trained length, while different heads in different layers can collectively process longer contexts.}
\model works efficiently in linear time, fits seamlessly with many LLMs that use relative positional encoding.
\model achieves 100\% accuracy at the \textbf{128k} length on passkey retrieval task, verifying \model's efficacy in extending the usable context window for existing models.
We release our code at \href{https://github.com/LuLuLuyi/LongHeads}{https://github.com/LuLuLuyi/LongHeads}.

\end{abstract}

\section{Introduction}

\begin{figure}[t!]
  \centering
  \includegraphics[width=\columnwidth]{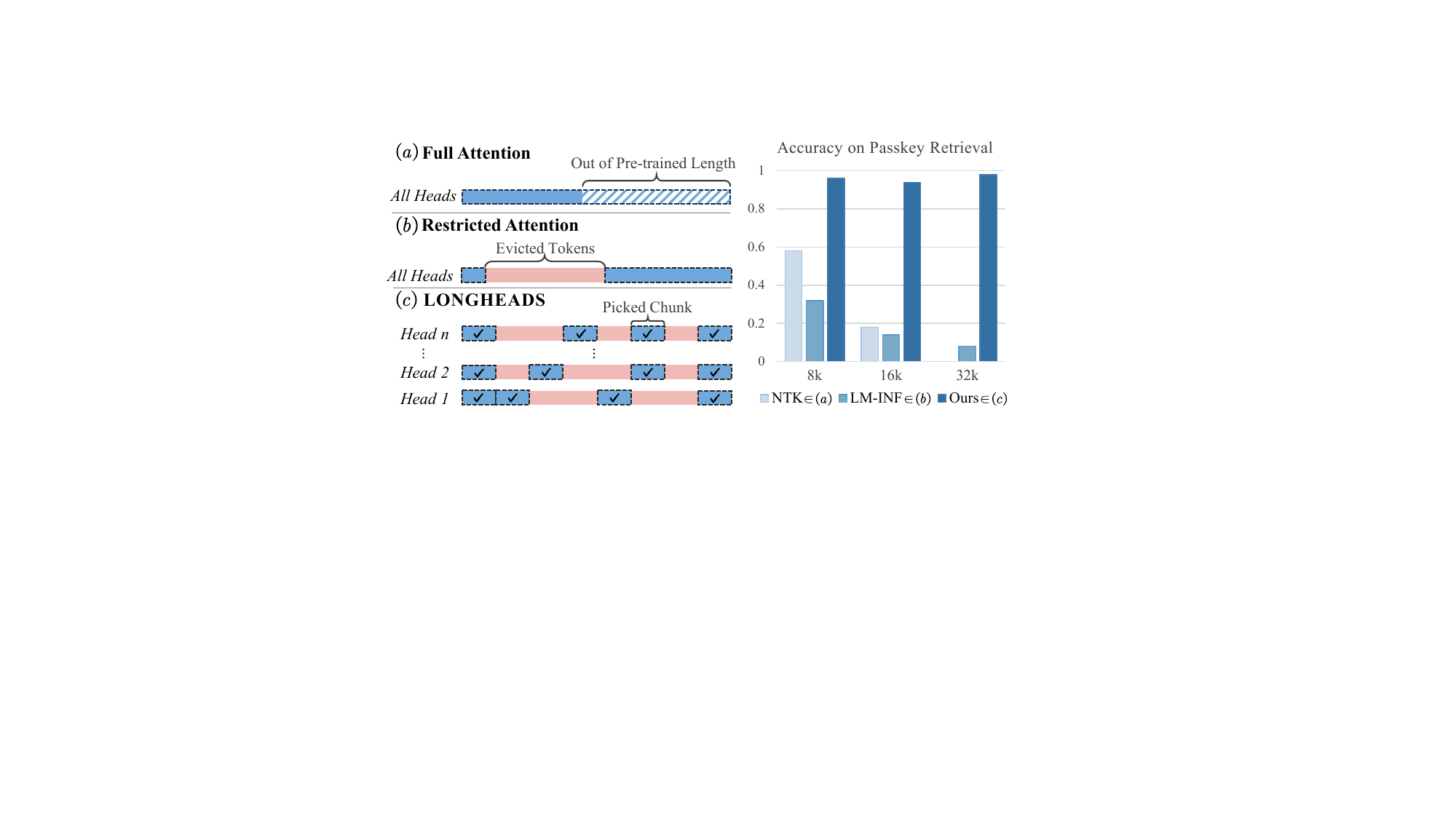}
  \caption{
  Left: Three types of long-context processors, (a) Attend all contexts \textbf{but} struggle with out-of-pre-trained length; (b) Attend local context to generate fluently \textbf{but} lose information; (c) Head attends short chunks \textbf{and} \textsc{Heads} attend \textsc{Long} context. Right: Accuracy of three specific methods 
  on passkey retrieval task.}
  \label{fig:ov_longhead}
  \vskip -0.15in
\end{figure}

LLMs are usually required to handle tasks with long contexts, such as in-context learning \citep{dong2023survey}, tool learning \citep{qin2023tool}, and retrieval-augmented generation \citep{gao2024retrievalaugmented}. 
However, enabling LLMs to process long contexts presents significant challenges. 
The OOD issue makes LLM struggle to process tokens beyond pre-trained length, and quadratic complexity of attention introduces considerable training and inference costs.
Although OOD issue could be addressed by 
zero-shot learning \citep{jin2024llm}, fine-tuning \citep{chen2023extending,peng2023yarn}, or re-training \citep{sun2022lengthextrapolatable,press2022train}, 
the required memory and computation still increases quadratically with context length, as shown in Figure~\ref{fig:ov_longhead}(a).

To alleviate these issues, recent works restrict the attention window to pre-trained length, which reduces the computation cost and avoids the processing of OOD tokens.
One direction is to exclude distant tokens (except for a few initial tokens, \citealp{han2023lminfinite, xiao2023efficient}) to restrict the attention window in-distribution, as shown in Figure~\ref{fig:ov_longhead}(b). 
However, 
these methods could result in losing critical information, degrading performance on downstream tasks.
The other way to constrain the attention window is to retrieve chunks of long sequences \citep{mohtashami2023landmark, zhang2024soaring}, but these approaches usually require special operations and continuous fine-tuning, which makes it difficult for existing LLMs to be directly applicable to long sequences.
In summary, improving the ability of LLMs to handle long contexts at a low cost is still challenging.

In this paper, we propose \textbf{\model}, a novel framework to enhance LLM’s long context ability without additional training. 
The key idea is to fully unlock the potential of multi-head attention.
We first utilize the nature of \textbf{different heads focus on different subspaces of the context, and each head can effectively process sequences within the pre-training length}. 
As shown in Figure \ref{fig:main} (c), we limit each head to selecting and attending to important contextual chunks within pre-trained length, rather than having each head attend to the entire sentence, thereby avoiding the OOD problem.
Furthermore, we leverage the model's inherent dot-product attention and propose a chunk selection strategy to find important chunks for each head. 
Drawing inspiration from the fact that \textbf{each head assigns different attention weights to tokens based on the inherent correlation between the query and the key representations}, 
we break the input into chunks and create chunk-level features for each block. It utilizes native token-level correlation to construct chunk-level queries and key representations, which allows each head to utilize its existing capabilities (dot-product attention) to select chunks based on the attention weights. 
In this way, each head effectively processes selected context chunks within the trained length, and all heads in all layers work together to handle longer contexts. Meanwhile, all operations are based on the intrinsic capabilities of multi-head attention, allowing \model to enhance LLMs without additional training.

To evaluate the effectiveness of \model, we employ LLaMA-2-7B-Base and LLaMA-2-7B-Chat as base models and evaluate on language modeling, synthetic retrieval task and long context benchmark.
\model achieving nearly 100\% accuracy across context lengths from 4k to 32k on the Passkey Retrieval task.
On LongBench, \model achieves the state-of-the-art (SOTA) performance among \emph{restricted attention} methods.
Compared with \emph{full attention} methods, \model achieves comparable performance on 16K test lengths and the best performance on 32K test lengths while enjoying linear computational cost. 
The experimental results demonstrate that \model enables the LLMs to directly generalize to longer sequences and achieve comparable or even superior performance compared to the methods that require continuous fine-tuning.

Our contributions can be summarized as follows: 
\begin{itemize}[leftmargin=*,topsep=1pt,itemsep=0.5pt]
\item We propose \model, a training-free inference framework that leverages the structural properties of attention heads to process long sequences efficiently and effectively.


\item We design a simple yet effective chunk selection strategy that can accurately select useful chunks and cover the full context.

\item Experiments demonstrate that \model is a SOTA restricted-attention-based long context processor and works efficiently in linear time, also with comparable performance to full-attention methods.

\end{itemize}

\section{Method}

In this section, 
we describe how the \model utilizes the inherent ability of multi-head attention to encode and generate long sequences without \emph{additional training}.

\subsection{Overview}

\begin{figure*}[t!]
  \centering
  \includegraphics[width=\textwidth]{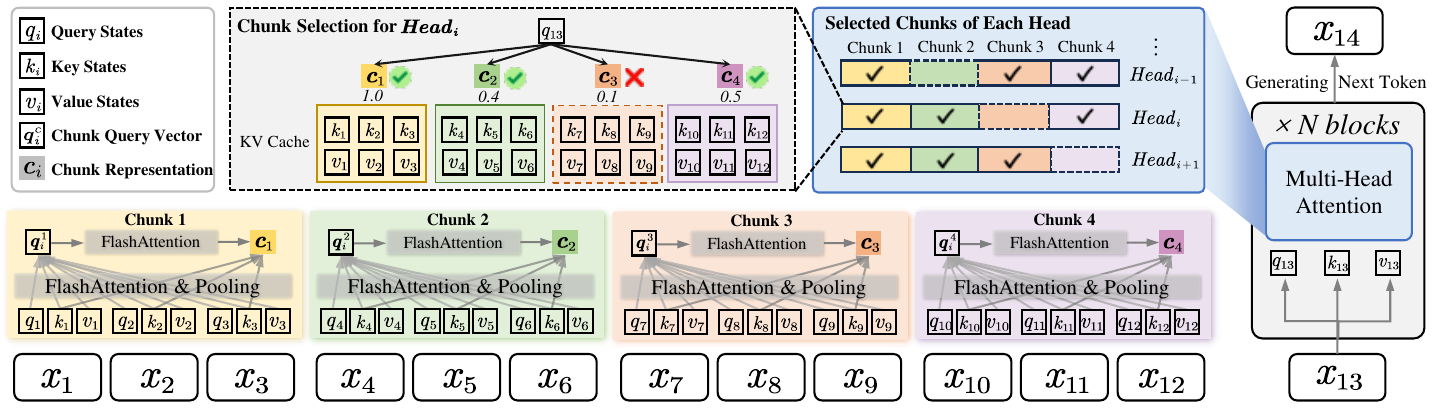}
  \caption{
  An overview of \model's inference, generating token $x_{14}$ in the current step. During inference, \model keeps the first chunk for stable computation, combined with the last chunk containing recent tokens.
  }
  \label{fig:main}
\end{figure*}

An overview of \model is shown in Figure~\ref{fig:main}. 
We break the text into chunks and calculate the chunk representations for each chunk. When generating token $x_{14}$, we pick the relevant $k$ chunks based on the current token's query vector and chunk representations.
In this way, each attention head of the \model selectively focuses on different text chunks according to its preference.
The tokens of attended chunks are then restructured, ensuring the subsequent causal attention always performed within the pre-trained length.

When encoding or generating an out-of-length token, 
a parameter-free chunk selection network picks the relevant $k$ chunks 
based on the current query vector and chunk representations.
Unpicked chunks can be approximated as having zero attention score 
(this usually holds under the sparsity of the attention mechanism), 
and do not need to be computed. 
This allows the attention matrix not to increase with length, 
significantly reducing the memory and computational cost of long contexts from $O(N^2)$ to $O(N)$.
Other works that restrict the scope of attention simply ignore distant tokens beyond a few initial tokens, 
even if they contain information worthy of attention.

In order to accurately select useful chunks, we utilize inherent similarity between token-level queries and token-level keys to construct chunk-level query and key representations.
Taking the 32K Passkey Retrieval experiment as an example,
the chunk containing the answer (i.e., the most valuable one) is the chunk with the highest selection score in 98\% of the cases without being trained.

\subsection{Chunk Representation}

Chunk representation is an indicator of whether the tokens in this chunk should be attended to. 
We obtain chunk representations in a training-free manner by utilizing the attention's intrinsic abilities.

Formally, given a long input sequence $X = (x_1,...,x_n)$, 
we segment it into chunks according to a predefined chunk size \(l\), 
then the input sequence can be denoted as $X=(C_1,...,C_m), m = \lceil \frac{n}{l} \rceil$.
We use attention's key states to generate chunk representation for each chunk due to the existing attention mechanism relies on query states.
There are numerous straightforward methods to obtain chunk representation, such as mean pooling of the key vectors of all tokens in the chunk. 
However, they have demonstrated suboptimal performance in preliminary experiments, particularly in selecting the correct chunks.
We hypothesize that this is attributed to the significance of individual tokens within a chunk vary substantially. 

To address the above problem, we should identify the tokens that can represent the entire chunk. 
For that purpose, we evaluate each token's significance to the chunk and perform scaled attention aggregation on all tokens' key states to obtain a representative chunk representation as follows:



\begin{equation}
\bm{c}_{i} = \text{flash-attention} \left(\bm{q}_{i}^{c},  \bm{K}_i, \bm{K}_i \right)
\end{equation}
where $\bm{c}_{i}\in \mathbb{R}^{m \times d}$ is the chunk representation, $\bm{K}_i \in \mathbb{R}^{l \times d}$ is the attention's all key states of chunk $C_i$, $\bm{q}_{i}^{c} \in \mathbb{R}^{m \times d}$ is a query vector to indicate which token's key state is suitable for representing the chunk representation. 
Next, we describe how to create the query vector. 


A good chunk query vector should be able to represent the chunk's full semantic information, i.e., the \emph{value} vector of all tokens in the entire chunk.
However, different tokens do not contribute equally to the semantic representation, e.g., content words hold a higher semantic weight, while function words contribute less.
Utilizing the inherent dot-product similarity between token-level query and key representations, we construct semantic weights for each token through a bidirectional self-attention aggregation.
From the perspective of message passing, semantically rich content words will transmit more of their information to other tokens, whereas function words transmit little.
Finally, the query vectors $\bm{q}_{i}^{c}$ that successfully summarize the complete semantics are obtained by mean-pooling of the aggregated representations, and can be formalized as follows. 
\begin{align}
\bm{O}_i &= \text{flash-attention}(\bm{Q}_i, \bm{K}_i, \bm{V}_i)  \nonumber\\
\bm{q}_{i}^{c} &= \text{mean}\left(\bm{O}_i \right),
\end{align}
where \(\bm{Q}_i\), \(\bm{K}_i\), and \(\bm{V}_i \in \mathbb{R}^{l \times d} \) are all query states, key states, and value states of chunk  \(C_i\) respectively.
Both \(\bm{K}_i\) and \(\bm{V}_i\) can be directly accessed from the KV cache, whereas \(\bm{Q}_i\) requires temporary storage during the calculation of the current chunk's representation and is released thereafter.

\subsection{Chunk Selection Strategy}

During the encoding or generation of the next token (denoted by $x_j$), 
we employ a query-aware chunk selection strategy, 
picking the $k$ most relevant chunks from those already generated.
Based on prior knowledge, there are two mandatory chunks.
One is aligning with \citet{xiao2023efficient}'s findings, 
acknowledging the essential role of the few start tokens of a sentence  
in preserving the stability of LLMs.
If the few start tokens are missing from the context, 
the pre-trained LLMs will completely lose their expressive ability 
(i.e., exhibit very high perplexity).
To ensure fluency, all attention heads uniformly select the first chunk (i.e., $C_1$) of the sentence. 
Otherwise, the LLM cannot handle downstream tasks (as demonstrated in the Ablation Study). 
The other is assigning the last chunk (i.e., $C_{-1}$) to all attention heads, in order to provide the model with the local information necessary for generation.

Next, we pick the remaining $k-2$ most relevant chunks for each attention head.
In the attention module of LLMs, the dot product score reflects the relevance of the context token to the current token. 
Inspired by it, 
we pick target chunks by the dot product similarity between the current token's query state $\bm{q}_j$ and the chunk representation $\bm{c}_i$.
\begin{equation}
P = \{C_1\} \cup \{C_i \mid \operatorname{rank}(\bm{q}_j\cdot\bm{c}_i) \le k-2\} \cup\{C_{-1}\},
\end{equation}
where $P$ is the final set of selected chunks, 
and the $\operatorname{rank}(\cdot)$ function outputs the rank of the current chunk's computed similarity among all candidates.
In this way, different attention heads across the layers naturally attend to different parts of the context, retrieving the important chunks for inference.

\paragraph{Position Remapping.}

\begin{figure}[h]
  \centering
  \includegraphics[width=\columnwidth]{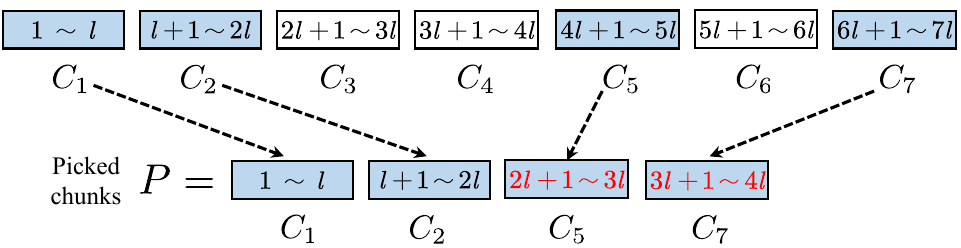}
  \caption{Demonstration of Position Remapping.}
  \label{fig:pe_remapping}
  \vskip -0.15in
\end{figure}

There are text chunks in the set $P$ that exceed the pre-training length, 
so the positional encoding of $P$ needs to be remapped.
The total length of the selected chunks is controlled to be within the pre-training length $L$, 
i.e., $k*l<L$.
Here, \textit{\model} restructures the picked chunks and concatenates them, 
while preserving the order of precedence.
In Figure \ref{fig:pe_remapping}, the current head attends to chunks $(1, 2, 7, 8)$ among the eight candidate chunks. 
The positions are assigned as $[1, 4l]$, in contrast to the original text positions, which would be  $[1, l] \cup [l\!+\!1, 2l] \cup  [6l\!+\!1, 7l] \cup [7l\!+\!1, 8l]$. 
Position remapping avoids the out-of-distribution problem encountered when extending the context even without further training.



\subsection{Inference with \model}

We separately describe the encoding of long inputs and the generation of long outputs during the inference.
Here we describe only the modified multi-head causal attention layer.

\paragraph{Computation and Memory in Encoding Phase.}
When the \textit{\model} receives long inputs, it first computes the representations of all chunks in parallel. 
This can be quickly achieved through two passes of \textit{flash-attention}, with the number of tokens involved in the attention equal to the chunk size (i.e., $l$=256, which is much smaller than the length of the input, e.g., $n$=16k). 
The second step is to select the $k$ most relevant chunks for each query based on chunk representations and to obtain their key and value representations, making the attention window equals to $k*l$=$w$ ( e.g., $w$=2k, which is also much smaller than $n$). 
Finally, length-restricted causal flash-attention is performed efficiently.

\paragraph{Computation and Memory in Generation Phase.}


During the generation process, \textit{\model} first performs chunk selection, then loads the Key-Value representations of the picked $k$ chunks for length-constrained causal attention.
When generating with very large inputs (e.g. 100K), the KV cache (except the chunk representations) can be offloaded to CPU to significantly reduce memory usage, and we only load the picked chunks into the GPU memory.
We always retain the query-key-value representations of recent tokens (not exceeding the chunk size) during the generation process. 
When the number of recent tokens equals the chunk size, we compute a chunk representation, similar to the encoding phase, and append it to the previous chunk representations.

Overall, the time complexity approximates an LLM with window attention O($w^2$) (window size $w$ is equal to $k*l$).
Memory usage of the decoding phase approximates O($n+w^2$), and can be further reduced to O($k*l+w^2$), avoiding a quadratic increase in costs with sequence length.

\section{Experiment}
We evaluate the proposed \model primarily using the LLaMA-2 \citep{touvron2023llama} considering its wide adoption and popularity.
The effectiveness of \model is evaluated on three kinds of tasks: language modeling, synthetic retrieval task and long context benchmark.
\subsection{Settings}
\paragraph{Implementation.}
Our method is applied to LLaMA-2-7B \emph{base} and \emph{chat} models for empirical studies. In our setup, we set the size of each chunk $l$ to be 256. 
During each inference step, we employ our chunk selection strategy to perform query-aware chunk selection. For each selection, we consistently choose the first chunk from the long text to facilitate normal generation by the model, and the last chunk to provide local context information. For all evaluation tasks, inference is conducted on a single NVIDIA A100 GPU. 

\paragraph{Baselines.}
The following types of baselines are chosen for comparison.
1) The method with full attention, including  ``Dynamic NTK'' interpolation (NTK, \citealp{emozillareddit}) and Position Interpolation (PI, \citealp{chen2023extending}). 
2) The method with restricted attention, including LM-Infinite \citep{han2023lminfinite} and Landmark-Attention \citep{mohtashami2023landmark}. 
The implementation details of baselines are in Appendix \ref{sec:baseline_implementation}.
 
\subsection{Long Context Language Modeling}
\begin{table}[t]
\centering
\small
\begin{tabular}{l@{\hspace{3pt}}r@{\hspace{3pt}}r@{\hspace{3pt}}r@{\hspace{10pt}}r@{\hspace{3pt}}r@{\hspace{3pt}}r}
\toprule
 & \multicolumn{3}{c}{\textbf{PG19}} & \multicolumn{3}{c}{\textbf{Proof-pile}} \\
\cmidrule(lr){2-4} \cmidrule(lr){5-7}
\textbf{Method} & 4k  & 16k & 32k  & 4k & 16k & 32k \\

\midrule
\rowcolor{gray!10} \multicolumn{7}{c}{\textit{\textbf{Full Attention}}} \\
PI-16K   & 7.42  & 6.72 & $>\!10^3$  & 2.98  & 2.61 & $>\!10^3$ \\
NTK  & 6.98  & 9.58 & 19.3  & 2.99  & 3.00 & 4.05 \\
\midrule
\rowcolor{gray!10} \multicolumn{7}{c}{\textit{\textbf{Restricted Attention}}} \\

LLaMA-2-7B   & 6.98  & $>\!10^3$ & $>\!10^3$  & 2.99 & $>\!10^3$ & $>\!10^3$ \\
LM-Infinite   & 6.98 & 7.33 & 7.75  & 2.99  & 2.96 & 3.10 \\
Landmark  & 10.03  & 10.13 & 10.14 & 4.98  & 4.86 & 4.92 \\
\model{}  & 6.98  & 8.15 & 8.41  & 2.99  & 3.26 & 3.42 \\
\bottomrule
\end{tabular}
\caption{Sliding window perplexity of different context window extension methods on PG19 and Proof-pile. 
\model extends the original LLaMA-2's context window length to 32k with 2k attention window.
}
\label{tab:perplexity-single-col}
\end{table}

\label{sec:long context Language Modeling}
The experiment on long context language modeling is performed with two datasets: PG19 \cite{rae2019compressive} and Proof-pile dataset \cite{azerbayev2023proofnet}. Details are shown in Appendix \ref{appendix:language_modeling_detail}.


The evaluation results are reported in Table \ref{tab:perplexity-single-col}.
Although the PPL of LLaMA-2-7B-Base model and PI remain low within the pre-training context length, it increases significantly when the context exceeds this window.
The NTK approach can maintain low PPL values for sequences up to 16k length, but PPL rises significantly at 32k context length. 
In contrast, \model, Landmark Attention and LM-infinite successfully maintain a low PPL score even at a sequence length of 32k. 


\subsection{Retrieval-Based Evaluation}
\label{subsec: pastkey}
We conduct experiments on the passkey retrieval task introduced by \citep{mohtashami2023landmark}. 
This task challenges a language model to accurately locate and retrieve a simple passkey (a five-digit random number) in a long text sequence. 
It tests whether a LLM can effectively attend to information across all positions of the input sequence. 
Following the design of \citet{mohtashami2023landmark}, the passkey is placed with various context lengths (ranging from 4k to 32k with 4k interval).
For each context length, we perform 50 tests with the passkey placed at a random position in the context.
\begin{table*}[ht]
\fontsize{15}{20}\selectfont
\setlength{\tabcolsep}{3pt}
\centering

\resizebox{\textwidth}{!}{
\begin{tabular}{lcccccccccccccccccc}
\toprule

\multirow{2}{*}{\textbf{Method}} & \multirow{2}{*}{\textbf{FT Tokens}} & \multicolumn{3}{c}{\textbf{Single-Doc QA}} & \multicolumn{3}{c}{\textbf{Multi-Doc QA}} & \multicolumn{3}{c}{\textbf{Summarization}} & \multicolumn{3}{c}{\textbf{Few-shot Learning}} & \multicolumn{2}{c}{\textbf{Synthetic}} & \multicolumn{2}{c}{\textbf{Code}} & \multirow{2}{*}{\textbf{Avg.}}\\
\cmidrule(lr){3-5}\cmidrule(lr){6-8}\cmidrule(lr){9-11}\cmidrule(lr){12-14}\cmidrule(lr){15-16}\cmidrule(lr){17-18}
 & & NQA & Qspr. & MulFi & HQA & WMQA & Musq. & GRpt & QMSM & MulN & TREC & TriQA & SMSM & PsgC & PsgR & Lcc & Repo  \\
 
\midrule 
\rowcolor{gray!10} \multicolumn{19}{c}{\textit{\textbf{Full Attention}}} \\
NTK & - & 16.47 & 29.62 & 31.42 & 31.31 & 28.75 & 10.20  & 22.70  & 17.65 & 6.31  & 64.67 & 77.36 & 37.95 & 3.99 & 5.12  & 65.64 & 52.97 & 31.38 \\
PI-16k & 0.85B & 21.37 & 31.78 & 36.67 & 37.56 & 27.47 & 15.98 & 13.55 & 20.69 & 1.18  & 63.00    & 89.24 & 25.64 & 5.67 & 11.33 & 67.05 & 56.02 & 32.76 \\

\midrule
\rowcolor{gray!10} \multicolumn{19}{c}{\textit{\textbf{Restricted Attention}}} \\
LM-Infinite & - & 14.34 & 20.75 & 26.18 & 20.37 & 20.08 & 5.87  & 16.70  & 7.01  & 2.28  & 54.67 & 76.69 & 15.64 & 4.30  & 7.00     & 62.90  & 52.74 & 25.47 \\
Landmark & 0.80B & 11.35 & 23.91 & 20.96 & 26.95 & 26.25 & 5.22  & 17.74 & 19.15 & \textbf{9.84}  & 42.67 & 80.73 & 35.45 & \textbf{5.73} & 7.00     & 59.74 & 42.76 & 27.22 \\
\textbf{\model}  & - & 14.51 & 21.58 & 30.32 & 30.07 & 25.28 & 9.15  & \textbf{24.74} & 20.26 & 6.30 & 55.00 & 83.26 & 34.27 & 2.45 & 9.39  & 65.01 & 50.65 & 30.14  \\

 \ \ \ \ w/ NTK init & - & 16.48 & 28.63 & 31.36 & 31.19 & \textbf{28.67} & 13.54 & 22.85 & 17.63 & 6.38 & \textbf{65.33} & 77.49 & \textbf{38.07} & 4.32 & 4.97  & 65.56 & \textbf{52.87} & 31.58 \\
 \ \ \ \ w/ PI init & 0.85B & \textbf{21.43} & \textbf{31.78} & \textbf{36.64} & \textbf{37.63} & 27.33 & \textbf{15.98} & 13.36 & \textbf{20.57} & 1.30 & 63.00 & \textbf{89.57} & 25.86 & 5.67 & \textbf{11.33} & \textbf{66.93} & 48.96 & \textbf{32.33} \\

\midrule 
\rowcolor{gray!10} \multicolumn{19}{c}{\textit{\textbf{Extend to 32k}}} \\
NTK & - & 5.74  & 29.05 & 31.39 & 28.98 & \textbf{27.03} & 9.34  & 22.00 & 15.13 & 5.40  & \textbf{64.67} & 48.34 & 34.50 & 3.89 & 4.85  & 57.54 & 45.29 & 27.07  \\
PI-16k & 0.85B & 8.43  & 30.15 & \textbf{35.20} & 29.47 & 24.72 & 1.74  & 13.23 & 12.59 & 1.30  & 55.00 & 66.15 & 19.16 & 5.42 & \textbf{11.33} & 33.21 & 27.21 & 23.39  \\
LM-Infinite & -  & 10.87 & 20.58 & 26.19 & 19.48 & 20.40 & \textbf{16.52} & 5.26  & 2.51  & 6.14  & 55.00 & \textbf{82.78} & 11.26 & 4.30 & 6.67  & \textbf{64.88} & \textbf{56.02} & 25.55 \\
Landmark & 0.80B  & \textbf{13.88} & 23.69 & 21.06 & 28.04 & 25.78 & 11.52 & 17.70 & 19.11 & \textbf{10.68} & 41.00 & 77.15 & \textbf{35.61} & \textbf{5.70} & 7.00  & 58.22 & 40.97 & 27.32 \\
\textbf{\model} & - & 13.38 & 21.81 & 30.33 & \textbf{29.59} & 24.90 & 11.48 & \textbf{27.43} & \textbf{19.87} & 6.07  & 55.00 & 81.15 & 33.56 & 2.79 & 10.06 & 63.75 & 47.97 & \textbf{29.95} \\

\bottomrule
\end{tabular}
}
\caption{The results of different methods based on the LLaMA-2-7B-Base model on \textbf{LongBench}. 
FT Tokens indicate the number of tokens used for continuous training.
The context window size for \model is 4k. 
}
\label{tab:longbench}
\end{table*}

\begin{figure}[t!]
  \centering
  \includegraphics[width=\columnwidth]{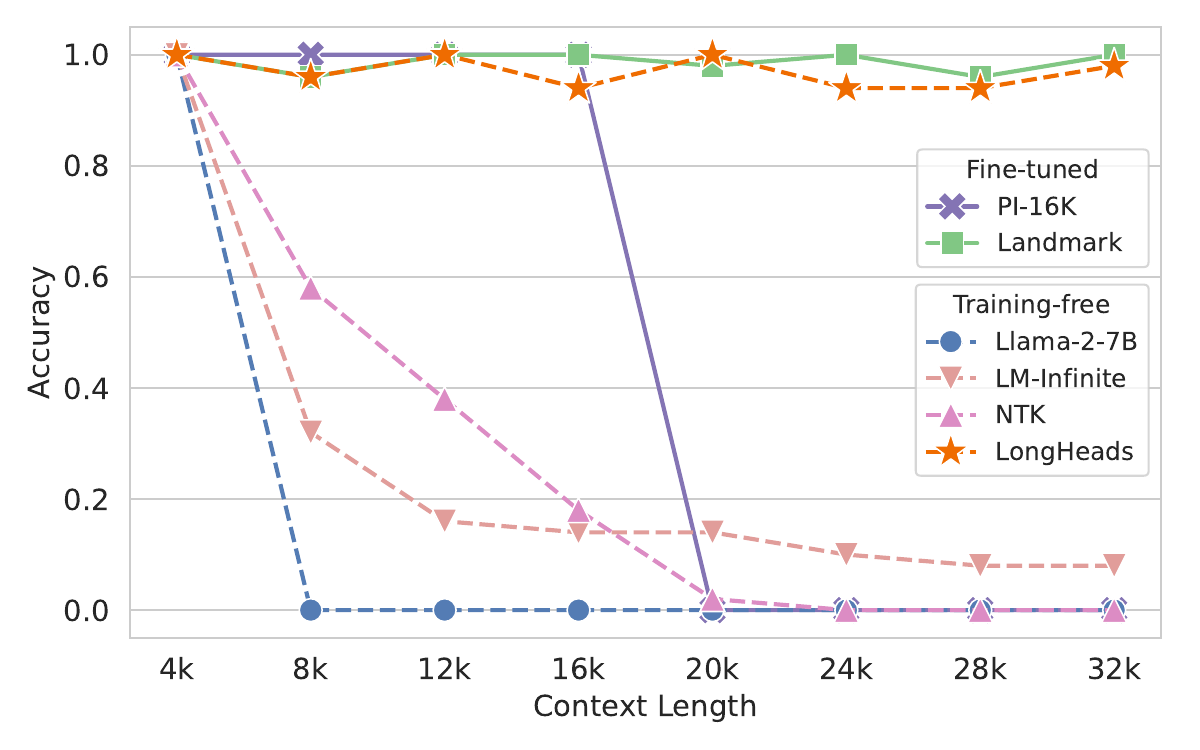}

  \caption{The evaluation of passkey retrieval task at different context lengths. \model achieves a comparable performance as Landmark Attention and outperforms other methods.}
  \label{fig:passkey_retrieval}
  \vskip -0.15in
\end{figure}

In Figure \ref{fig:passkey_retrieval}, we can see that all the models can output the passkey within the pretrained length. 
The base model completely fails at the extended length.
The NTK and LM-Infinite induce a significant drop in accuracy for models at lengths surpassing 6k tokens, with accuracy falling below 20\% when token lengths exceed 16k. 
LM-Infinite can only access 10\% passkey with its local window, despite having low PPL at 32k length.
Conversely, Landmark Attention and \model consistently retrieve with nearly 100\% accuracy regardless of sequence length.

We further test \model at 128k length after offloading KV cache to CPU, the results are shown in Appendix \ref{sec:passkey_128k}.
We note that \model uses only 2k attention window achieving 100\% accuracy at the 128k length without training.


\subsection{Long Context Benchmark Evaluation}


Language modeling tasks have proven to be insufficient metrics for ensuring success in downstream tasks \citep{sun2021longrange}, while synthetic password retrieval tasks often do not align with real-world scenarios.
It is significant to conduct real downstream task evaluations to more comprehensively reflect the model's long sequence capabilities. We opt LongBench \citep{bai2023longbench} for downstream NLP task evaluation, the details are shown in Appendix \ref{appendix:longcontext_benchmark_detail}.
The results are listed in Table \ref{tab:longbench}. 
We also conduct experiments on LLaMA-2-7B-Chat model, and the results are shown in Appendix \ref{appendix:longbench_results}.

\paragraph{Comparison with Restricted Attention Methods.}
\model{} surpasses the current methods with restricted attention. 
Specifically, \model{} performs better than the method with the sliding window mechanism on LongBench (+4.67 vs. LM-Infinite).
Compared to the method with chunking strategy (i.e., Landmark Attention), \model{} exceeds the average score by 2.92 on LongBench without additional training.
This indicates that the chunk selection strategy in \model{} can accurately supplement LLMs with relevant contextual information, enabling efficient and effective understanding on long sequences.

\paragraph{Comparison with Full Attention Methods.}
Full attention methods can increase the maximum sequence length of LLMs but also raise computational and memory costs.
\model{} can be augmented with PI or NTK methods during the encoding phase, achieving comparable or even better results with a shorter window size, significantly reducing computational overhead.
This suggests that \model{} has the potential for scalability, and can be strengthened with a stronger base model.

\paragraph{Performance when extending to 32k Context window.}
A desirable attribute for RoPE-extension methods is that the models should maintain their performance when directly extending to a longer context window. 
When extending to 32k context windows, PI and NTK methods struggle with the out-of-demonstration issue and tend to compromise model performance. 
In contrast, \model{} maintains its performance and outperforms all the baseline methods.
It successfully extend LLaMA-2-7B-Base from a 4K length to 8 times its length, demonstrating that \model{} can easily generalize to a longer context window.


\begin{figure*}[ht]
  \centering
  \includegraphics[width=\textwidth]{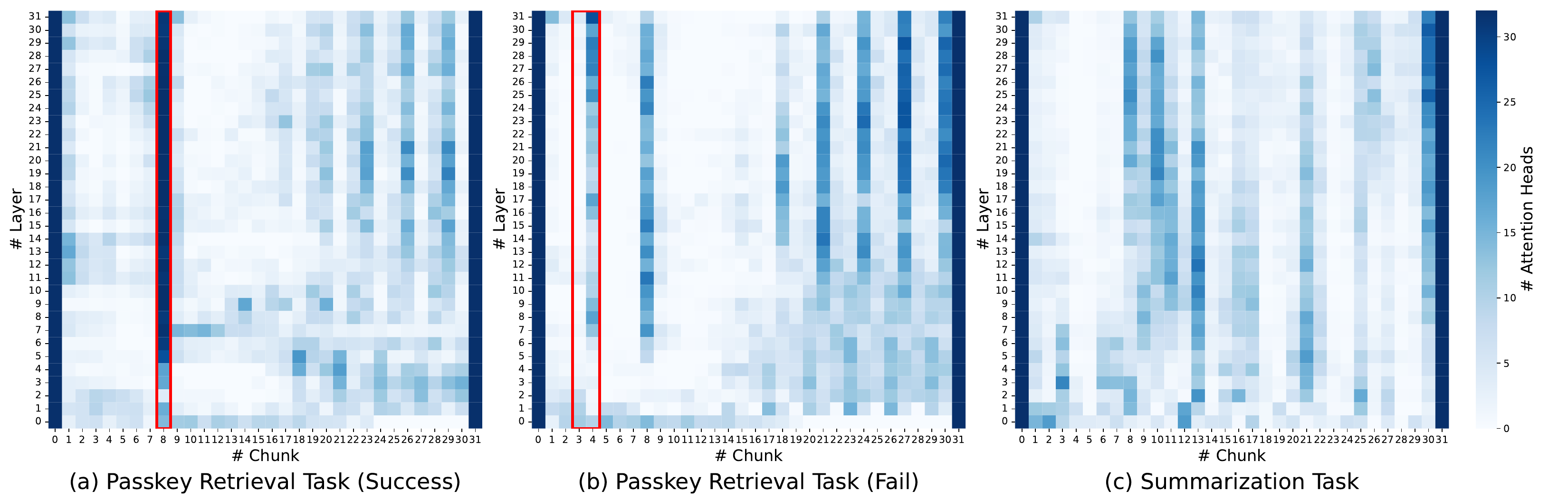}

  \caption{
 Visualization of chunks selected by different attention heads at each layer represented by color blocks.
  For the passkey retrieval task, the chunk containing the passkey is delineated with a red border. 
  For the failed example, the red border encompasses two chunks due to the passkey-containing sentence coincidentally spanning two chunks.
  }
  \label{fig:heatmap}
\end{figure*}

\section{Discussion}
\subsection{Analysis}
\label{sec:Analysis}

In this section, we explore how different attention heads handle long contexts and whether they find important information.
We set \model's attention window to 2048 and analyzed its performance on passkey retrieval and summary tasks.
We visualize the tests for both tasks in Figure \ref{fig:heatmap} and show the statistical results in Table \ref{tab:Analysis}. The details of analytical experiments are in Appendix \ref{appendix:analysis_detail}.

\begin{table}[t]
\centering
\small
\begin{tabular}{ccccc}
\toprule
 \textbf{Input}   &  \textbf{Cover}  &  \multirow{2}{*}{\textbf{Uniformity}} & \multicolumn{2}{c}{\textbf{Hit Rate}} \\
\cmidrule(lr){4-5}
\textbf{Length} & \textbf{Rate} & & Top 1 & Top 5 \\
\midrule
\rowcolor{gray!10} \multicolumn{5}{c}{\textit{\textbf{Passkey Retrieval}}} \\
 4k & 100 & 0.52 & 0.55 & 0.96 \\
 8k & 100 & 0.52 & 0.89 & 0.96 \\
 16k & 99.2 & 0.60 & 0.99 & 1.00 \\
 32k & 82.0 & 0.76 & 0.98 & 0.98 \\
\midrule
\rowcolor{gray!10} \multicolumn{5}{c}{\textit{\textbf{Summary}}} \\
 4k & 100 & 0.31 & / & / \\
 8k & 100 & 0.44 & / & / \\
 16k & 100 & 0.49 & / & / \\
 32k & 100 & 0.57 & / & / \\
\bottomrule
\end{tabular}
\caption{Statistical results with different sequence lengths. 
Cover Rate is defined as the percentage of selected chunks out of the total number of chunks. 
Uniformity of the distribution of chunk selection is evaluated by the Gini coefficient, with lower values indicating a more uniform distribution. Hit Rate means the probability that the top-1 and top-5 selected chunks contain the correct answer in the past key retrieval task.
}
\label{tab:Analysis}
\vskip -0.15in
\end{table}

\paragraph{Attention heads focus on important parts in context.}
On the passkey retrieval task, shown in Figure \ref{fig:heatmap}(a), all attention heads focused on the same chunk containing the answer and predicted it accurately.
Even when the passkey is not successfully predicted in Figure \ref{fig:heatmap}(b), the chunks containing the answer are still selected by multiple heads.
In contrast, on the summary task in Figure \ref{fig:heatmap}(c),  the attention heads spread their focus more evenly to summarize the entire information.
Similarly, Table \ref{tab:Analysis} reveals a lower uniformity score for the summary task compared to the passkey retrieval task.
These findings suggest that our chunk selection strategy results in a more uniform distribution of selections in the summary task, while the distribution in the passkey retrieval task is more concentrated.
We attribute this to the specificity of chunks required for the passkey retrieval task, 
whereas the summary task necessitates various parts of the text to formulate a comprehensive answer.
Moreover, the probability of the top 5 selected chunks containing the answer is almost 100\% across all test lengths in Table \ref{tab:Analysis}. 
These results suggest that our chunk selection strategy adaptively fits the characteristics of different tasks, and allows different attention heads to concentrate on task-related content.

\paragraph{Attention heads can handle long sequences in a short window.}
In Figure \ref{fig:heatmap}, the lower layer attention heads focus on the more dispersed text in both tasks, while the upper layer attention heads focus more on specific chunks. 
We speculate that different attention heads naturally focus on different parts of the information in the text at lower layers, collecting and aggregating the entire long document information in a short length, while the upper layer attention heads are responsible for processing the aggregated information, mainly focusing on the chunks needed to complete the task.
In Table \ref{tab:Analysis}, the Cover Rate is 100\% in most cases.
Given that different heads in each layer can select varying chunks, the maximum theoretical length accessible by \model is $|P|\times \operatorname{{n\_heads}}\times \operatorname{{n\_layers}}$ (e.g., the maximum length for LLaMA-2-7B with 4k attention window is 512k).
These observations demonstrate that we have successfully utilized a limited attention window to capture almost all information from the entire long document.

\subsection{Ablation Study}

We conduct ablation experiments to investigate the influence of chunk selection strategy, attention heads flexibility, number of chunks $K$, and chunk size $l$.
The ablation study is constructed on LongBench and the results are presented in Table \ref{tab:ablation}.
\begin{table}[t]
\centering
\resizebox{0.4\textwidth}{!}{%
\begin{tabular}{lc}
\toprule
\textbf{Method Setting} & \textbf{LongBench Avg.} \\ 
\midrule
\model & \textbf{30.14}\\
\hdashline
\ \ \ \   - Random Selection  & 28.77 \\
\ \ \ \   - Last K Selection  & 26.22 \\
\ \ \ \  - w/o First Selection & 14.06 \\
\hdashline
\ \ \ \   - Fix Head  & 29.46 \\
\ \ \ \   - Fix Layer  & 28.78 \\
\ \ \ \  - Fix Head \& Layer & 28.72 \\
\hdashline
\ \ \ \   - Number of Chunks $K=8$ & 29.09 \\
\ \ \ \  - Number of Chunks $K=4$ & 26.64 \\
\hdashline
\ \ \ \   - Chunk Size $l=512$ & 29.95 \\
\ \ \ \  - Chunk Size $l=128$ & 29.35 \\

\midrule
\end{tabular}%
}
\caption{Ablation study on \textbf{LongBench}, by default $l=256$, $K=16$, and Top K Selection.}
\label{tab:ablation}
\vspace{-1.5mm}
\end{table}
\paragraph{Effect of Chunk Selection Strategy.}
We find that the performance when selecting the highest-scoring chunks significantly surpasses that of the lowest-scoring (Last K) chunks, and even Random Selection yields better results than Last K Selection. 
We also observe a significant performance degradation when the first chunk is not preserved. 
This is because the absence of the first chunk results in the model's output distribution collapsing directly.
Our findings are consistent with StreamingLLM \citep{xiao2023efficient} and LM-Infinite \citep{han2023lminfinite}.

\paragraph{Effect of Heads Flexibility.}
When the flexibility of attention heads is constrained, the model's performance is compromised to varying degrees (-0.68 Fix Head, -1.36 Fix Layer, -1.42 Fix Head\&Layer).
This demonstrates that within the \model framework, the collaboration of different attention heads in each layer plays a crucial role.

\paragraph{Effect of Number of Chunks \& Chunk Size.}
Increasing the number of chunks in a text may provide more information, but the benefits show a diminishing return. This indicates that four chunks provide enough information to ensure performance, and eight chunks are already adequate to access the entire sequence's information with chunk selection strategy,
Different chunk sizes do not lead to a significant impact on the results, indicating larger or smaller chunk sizes are feasible for \model.

\section{Related Work}

\paragraph{Expanding Positional Encoding (PE).}
Context extension studies typically target the popular RoPE encoding, aiming to scale unseen PE into the space of positions seen during pre-training.
\citet{chen2023extending}, and concurrently \citet{kaiokendev9444} 
discovered that interpolating the position indices within the pre-trained limit works well with the help of a small amount 
(a few billion, \citealp{chen2023extending}) of fine-tuning. 
However, 
position interpolation (PI) equally stretches all dimensions of RoPE, 
neglecting the variations in frequency. 
As an alternative, 
\citet{bloc97} 
proposed the ``NTK-aware'' interpolation by taking the loss of high-frequency components into account.
Subsequently, 
\citet{emozillareddit} proposed the ``Dynamic NTK'' interpolation method, 
which performs well without the need for fine-tuning.
\citet{bloc972} 
introduced the ``NTK-by-parts'' interpolation method, 
which performs the best when fine-tuned on a small amount of longer-context data.
\citet{peng2023yarn} 
proposed YaRN,
an improved method to efficiently extend the context window by fine-tuning on less than 0.1\% of the original pre-training data.
This work directly modifies the PE to expand to a theoretically infinite context length. 
In contrast, our method does not require modifying the PE, 
and only a finite chunk participates in the attention calculation during generation, 
which improves inference efficiency and reduces memory usage.

\paragraph{Restricted Attention.}
In addition, the global causal attention could be restricted to local attention, 
thus avoiding exceeding the pre-trained position length.
ReRoPE \citep{rerope2023} truncates all context lengths to the max length during pretraining.
LM-Infinite \citep{han2023lminfinite} restricted the global attention window into a chevron-shaped window, 
retaining only a few tokens from the beginning of the text and a local window.
\citet{mohtashami2023landmark} insert a learnable landmark token after each text fragment with a fixed length, and use these landmarks to retrieve relevant fragments.
\citet{zhang2024soaring} similarly insert a learnable beacon token and use its representation to summarise the corresponding whole fragment.
Although restricted attention offers advantages in terms of memory usage and inference speed, 
they risk losing valuable context information. 
Existing methods employ local windows that are either fixed or selected through fine-tuning. 
In our approach, local windows are flexibly composed of chunks from the context and do not rely on additional fine-tuning.

\section{Conclusion}
We present \model, a novel, training-free framework for efficiently processing long contexts in pre-trained LLMs. 
Utilizing the intrinsic capabilities of attention heads, \model smartly segments and assigns long text to relevant heads, streamlining the handling of extended sequences without extra computational load. 
Experiment results validate \model's superiority in restricted attention setups and its competitive edge against full attention methods when applied to the LongBench suite. 
Our approach paves the way for performance breakthroughs in long context LLM operations, leveraging existing model structures to unlock new potential without further training.

\section*{Limitations}
We summarize the limitations of our method as follows:
(1) Splitting the text into chunks may disrupt the continuity of the content. When the correct answer is in the middle of two chunks, this kind of splitting can affect the performance of downstream tasks.
(2) The theoretical maximum length accessible by \model is confined to $|P|\times \operatorname{n\_heads}\times \operatorname{n\_layers}$. \model cannot fully access inputs that surpass this threshold. However, \model can still perform well on long document tasks by selecting important parts from the context.
(3) The success of \model in downstream tasks depends on the non-parametric chunk selection function. For complex comprehension tasks, the effectiveness of the selection function may be affected.

\bibliography{custom}

\appendix

\section{Baseline Implementation Details}
\label{sec:baseline_implementation}
We conduct experiments on 4 methods as our baselines.We illustrate the details of each baseline as follows:

For NTK, we set the scale factor of NTK to 2.0 for base model and 1.0 for chat model. 
For LM-Infinite, we set the number of preserved initial tokens to 10 and the local window at the end to 4096 tokens. In the context of training-free methods, we did not evaluate StreamingLLM \citep{xiao2023efficient} as their framework does not support inputs exceeding 4K tokens, and their method is similar to LM-Infinite.
For Position Interpolation method performed on 8K and 16K context, we use the Redpajama \citep{redpajama} dataset for training. Following \citep{chen2023longlora}, we set the per-device batch size as 1 and gradient accumulation steps as 8, which means that the global batch size equals 64, using 8 GPUs. We train the models for 1000 steps. 
For Landmark-Attention, we adopted their configuration settings for consistency. We finetune LLaMA-2-7B Base model for 15000 steps using their method. We fine-tune the model with context length 512 on Redpajama dataset.

\section{Passkey Retrieval Evaluation on 128k context}
\label{sec:passkey_128k}
\begin{figure}[t!]
  \centering
  \includegraphics[width=\columnwidth]{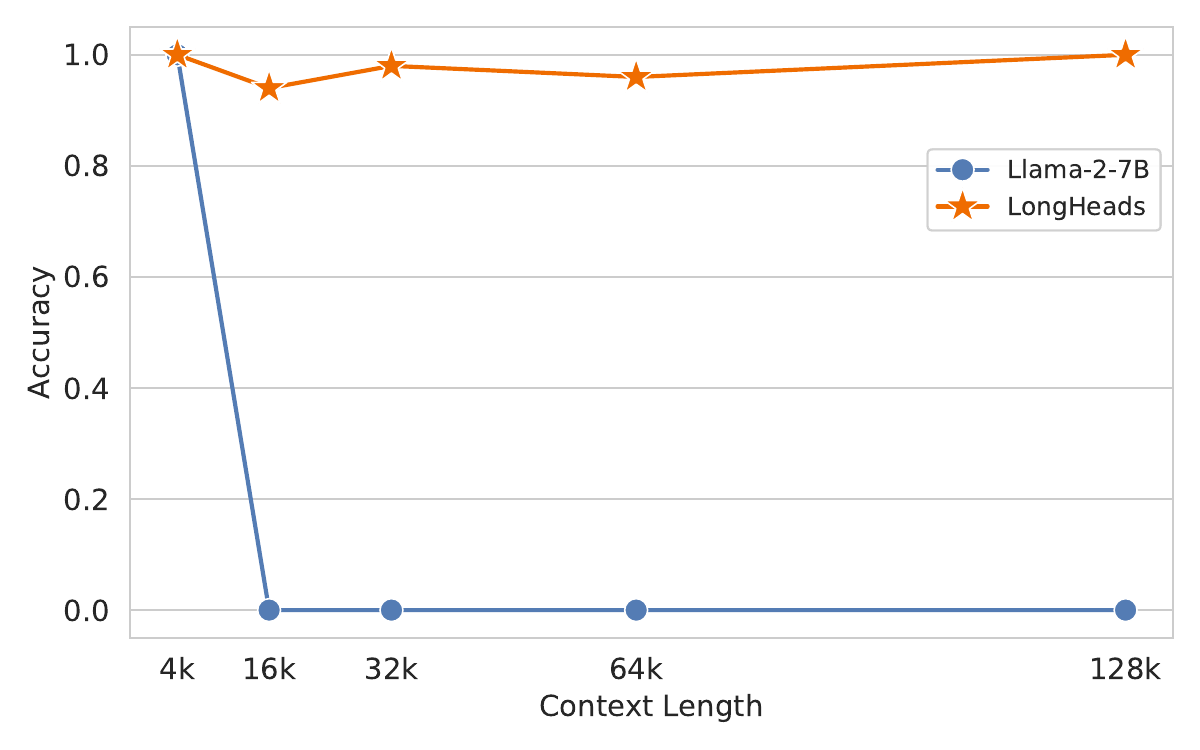}

  \caption{The evaluation of passkey retrieval task from 4k to 128k.}
  \label{fig:passkey_128k}
  \vskip -0.15in
\end{figure}
We further extend LLaMA-2-7b to 128k with LongHeads without additional training. \model achieves 100\% accuracy at 128k length on passkey retrieval task, the results are shown in Figure \ref{fig:passkey_128k}. 
After offloading the KV cache to CPU, peak GPU memory usage is 26.51GB and 44.48 GB when inference with 64k and 128k context.
\begin{table*}[!ht]
\fontsize{15}{20}\selectfont
\setlength{\tabcolsep}{3pt}
\centering

\resizebox{\textwidth}{!}{
\begin{tabular}{lcccccccccccccccccc}
\toprule

\multirow{2}{*}{\textbf{Method}} & \multirow{2}{*}{\textbf{FT Tokens}} & \multicolumn{3}{c}{\textbf{Single-Doc QA}} & \multicolumn{3}{c}{\textbf{Multi-Doc QA}} & \multicolumn{3}{c}{\textbf{Summarization}} & \multicolumn{3}{c}{\textbf{Few-shot Learning}} & \multicolumn{2}{c}{\textbf{Synthetic}} & \multicolumn{2}{c}{\textbf{Code}} & \multirow{2}{*}{\textbf{Avg.}}\\
\cmidrule(lr){3-5}\cmidrule(lr){6-8}\cmidrule(lr){9-11}\cmidrule(lr){12-14}\cmidrule(lr){15-16}\cmidrule(lr){17-18}
 & & NQA & Qspr. & MulFi & HQA & WMQA & Musq. & GRpt & QMSM & MulN & TREC & TriQA & SMSM & PsgC & PsgR & Lcc & Repo  \\
 
\midrule 
\rowcolor{gray!10} \multicolumn{19}{c}{\textit{\textbf{Chat Model}}} \\

LM-Infinite & -  & 0.00     & 18.57 & 25.33 & 9.87  & 11.73 & 0.48  & 11.30  & 2.99  & 8.72  & 32.50  & 29.22 & 13.82 & 5.61 & 5.20   & 34.19 & 24.55 & 14.63 \\
NTK & - & 15.18 & \textbf{30.89} & 36.14 & 35.10  & 25.79 & \textbf{13.53} & \textbf{31.48} & 20.21 & 23.86 & \textbf{61.67} & \textbf{80.94} & \textbf{39.43} & \textbf{7.40}  & 13.33 & 48.96 & 42.45 & \textbf{32.90}  \\
\textbf{\model} & - & 11.61 & 22.98 & 23.76 & 31.28 & 24.10  & 8.87  & 25.36 & 20.24 & 16.18 & 50.67 & 79.98 & 36.74 & 6.39 & 9.67  & \textbf{53.85} & \textbf{44.22} & 29.12 \\
 \ \ \ \ w/ NTK init & - & \textbf{16.87} & 30.32 & \textbf{38.59} & \textbf{36.04} & \textbf{26.72} & 10.21 & 31.28 & \textbf{20.91} & \textbf{24.46} & 55.67 & 76.72 & 39.07 & 6.07 & \textbf{14.67} & 49.97 & 40.27 & 32.37 \\

\bottomrule
\end{tabular}
}
\caption{The results of different methods based on the LLaMA-2-7B-Chat model on \textbf{LongBench}.}
\label{tab:longbench_appendix}
\end{table*}
\section{Evaluation Details}
\subsection{Language Modeling Evaluation Details}
\label{appendix:language_modeling_detail}
We evaluate the long context language modeling performance on the book corpus dataset PG19 \cite{rae2019compressive} and the cleaned Arxiv Math proof-pile dataset \cite{azerbayev2023proofnet}.
For both datasets, a subset of one hundred instances from the test corpus is utilized to gauge language modeling proficiency.
Following \cite{press2022train}, we evaluate perplexity by using a sliding window approach with S = 256.


\subsection{Long Context Benchmark Evaluation Details}
\label{appendix:longcontext_benchmark_detail}
Following \citet{jin2024llm,zhang2024soaring},
we opt Longbench \citep{bai2023longbench} for downstream NLP task evaluation, including Single-Document Question Answering (QA), Multi-Document QA, Summarization, Few-shot Learning, and Code Completion. To ensure a more balanced and rational evaluation of the model's long-text capabilities, we employed tasks from LongBench-E to replace the corresponding tasks in Longbench for our testing.
We follow LongBench \citep{bai2023longbench} to evaluate the models on 16k context window sizes by truncating the prompt from the middle when the task length exceeds a designated context window size.

\section{Analysis Experiments Details}
\label{appendix:analysis_detail}
We conduct analytical experiments on the tasks of passkey retrieval and summary.
For the passkey retrieval task, we compiled statistics for the results with sequence lengths of 4k, 8k, 16k, and 32k, as mentioned in Section \ref{subsec: pastkey}. 
Regarding the summary task, we select the government report dataset from the LongBench, from which we chose 5 samples each for lengths of 4k, 8k, 16k, and 32k for statistical analysis.

\section{More Results on LongBench}
\label{appendix:longbench_results}
Tabel \ref{tab:longbench_appendix} shows that \model{} also has strong performance on LLaMA2-7b-Chat models. When encoding is enhanced with NTK, \model{} is able to achieve comparable performance to the full attention method.



\end{document}